\documentclass[conference]{IEEEtran}

\usepackage[margin=0.795in]{geometry}

\usepackage{graphicx}

\usepackage{adjustbox}
\usepackage{multirow}

\author{
        Ali Atghaei,\IEEEmembership{}
        Soroush Ziaeinejad,~\IEEEmembership{}and~Mohammad Rahmati~\IEEEmembership{Member,~IEEE,}% <-this % stops a space

\IEEEauthorblockA{\\Department of Computer Engineering and Information Technology\\
Amirkabir University of Technology (Tehran Polytechnic), Tehran, Iran\\
Email: \{atghaei, ziaeinejad, rahmati\}@aut.ac.ir}

}

\title{Abnormal Event Detection in Urban Surveillance Videos Using GAN and Transfer Learning}
\usepackage{cite}
\usepackage{amsmath,bm}
\usepackage{mathtools}
\usepackage{array}
\usepackage{tabularx}
\usepackage[T1]{fontenc}
\usepackage{makecell}
\usepackage{tikz}
\usepackage{fancyhdr}

\date{}

\usepackage{fancyhdr}
\begin{document}
\maketitle

\thispagestyle{fancy}

\begin{abstract}
%\boldmath
Abnormal event detection (AED) in urban surveillance videos has multiple challenges. Unlike other computer vision problems, the AED is not solely dependent on the content of frames. It also depends on the appearance of the objects and their movements in the scene. Various methods have been proposed to address the AED problem. Among those, deep learning--based methods show the best results. This paper is based on deep learning methods and provides an effective way to detect and locate abnormal events in videos by handling spatio-temporal data. This paper uses generative adversarial networks (GANs) and performs transfer learning algorithms on pre-trained convolutional neural network (CNN) which result in an accurate and efficient model. The efficiency of the model is further improved by processing the optical-flow information of the video. This paper runs experiments on two benchmark datasets for AED problem ($\text{UCSD Peds1}$ and $\text{UCSD Peds2}$) and compares the results with other previous methods. The comparisons are based on various criteria such as area under curve (AUC) and true positive rate (TPR). Experimental results show that the proposed method can effectively detect and locate abnormal events in crowd scenes.
\end{abstract}


\section{Introduction}
Using smart surveillance cameras has recently become very popular. Compared to human monitoring systems, these smart cameras have more consistent behavior and can provide higher accuracy and quicker response. The higher efficiency of these systems has attracted researchers and developers active in developing automated surveillance systems~\cite{chalapathy2019deep}. By considering the widespread use of automated surveillance systems in different applications, it is expected that computer vision--based systems will be able to automatically process a large amount of visual information. Abnormal Event Detection (AED) in videos is one of the most popular computer vision topics. Because of the nature and subjective definitions of ‘abnormality’ (or ‘anomaly’), AED is a very challenging and content-dependent problem~\cite{sabokrou2018deep}. with its major problems addressed, AED can be used for a wide variety of applications such as crowd analysis, subway stations and urban pathways surveillance, summarization of surveillance videos, and smart home monitoring. In the case of using AED for crowd analysis, this system is expected to understand the crowd behavior in a public place for a period of time and inform human agents if it observes an unusual event in the video.

An event is considered abnormal if it is unlikely or unexpected to occur. In statistics, ‘anomaly’ is defined as an unusual behavior in a distribution or an outlier data point in a data space. Anomaly detection systems are trained with a normal dataset and construe an  outlier as abnormal behavior. These systems usually try to jointly detect and locate abnormal behavior. Locating means detecting the location of an abnormal event by showing the abnormal pixels of each frame. For AED to have high accuracy and quick response, an effective data representation method is required. Data space in AED is spatio-temporal with both appearance and movement involved in the process. Researchers use various methods to locate abnormal parts of a video file. The most popular method is gridding, which splits a sequence of frames to smaller fixed-size 3D patches by applying a fixed grid on the frames~\cite{xu2017detecting}.

%There are multiple benchmark datasets for AED that are commonly used to evaluate any new method and compare it with the state of the art.

Crowd scene analysis has multiple challenges such as occlusion, shadowing, and overlapping of moving objects. Different algorithms have been proposed to overcome these challenge but all of them have their advantages and disadvantages. There are several methods to understand the movement information of scene elements. These methods are able to precisely model the direction and speed of each individual object. However, these methods are usually very time consuming. Besides, the perspective distortion of urban surveillance videos adds to the complexity of the problem as it causes different scale and movement patterns based on object locations and camera position. Because of different lighting situations and subtle difference between normal and abnormal cases, an accurate discriminative model is needed to detect abnormal patches and frames. Various machine learning models such as deep neural networks require a large amount of labeled data. However, AED is an unsupervised learning problem for which gathering a large amount of labeled data is a time consuming and arduous task~\cite{sabokrou2018deep}.

Several methods are proposed for AED problem. Trajectory-based methods give a highly accurate model by using the movement of scene entities. Recently, some methods such as Histogram of Gradients (HoG) or Histogram of Optical Flows (HOF) are used to model spatio-temporal properties of the videos.  In these methods, an entity is considered abnormal if the model has never observed its movement pattern before. However, these methods are not efficient for crowd scenes because of their high time complexity and also the problem of moving object occlusion~\cite{jiang2011anomalous},~\cite{wu2010chaotic}. Researchers have published several papers to address the AED problem. A considerable portion of these published research is evaluated with $\text{UCSD Peds}1$ and $\text{UCSD Peds2}$ datasets ~\cite{mahadevan2010anomaly}.

Researchers had used traditional machine learning methods before creating deep learning models. Although these methods are still in use, they have become limited to specific applications. An HMM model using combined dynamic texture as the feature set~\cite{mahadevan2010anomaly}, social force method using spatio-temporal data filtering~\cite{mehran2009abnormal}, sparse representation method~\cite{cong2013abnormal}, optical flow clustering method~\cite{saligrama2012video}, bag-of-visual-word model to represent images~\cite{javan2013online}, and a GMM model using 3D gradient images~\cite{feng2017learning} are examples of these traditional models.

 Another class of machine learning models is deep learning networks. Recent research activities commonly use these models to solve complex problems. Researchers also use deep models to address the AED problem. A one-class SVM model using optical flow features which extracted by an auto-encoder network is proposed in~\cite{xu2017detecting}, and a one-class SVM model using feature set which is extracted by a pre-trained deep model is proposed in~\cite{sabokrou2018deep}.

%{\color{red} [[WILL BE REMOVED]] Convolutional neural network (CNN) is one of the most popular neural networks for image and video processing. This network is highly efficient in solving image processing problems. However, it has two major drawbacks:

%\begin{itemize}
%\item CNN usually has a high time complexity in processing visual data. Additional algorithms are required to reduce this complexity.
%\item CNN requires a large amount of labeled data to be trained properly using a supervised training algorithm. It is virtually impossible to obtain a large abnormal events dataset.
%\end{itemize}}

This paper uses a new method to address the AED problem based on two essential steps: Spatio-temporal features processing and motion analysis using optical flow images. Fig.~\ref{fig1} shows the block diagram of the learning phase of the proposed method which has three main parts: pre-processing, training the GAN, and analyzing the optical flow images. The testing phase includes a pre-processing step which is then followed by a spatio-temporal representation. The representation is then calculated and fed to the trained discriminator network as an input. In the final step, abnormal patches of all the frames are determined and an output image is created to show the appearance-motion abnormalities of each frame.

The rest of this paper is organized as follows: Section~2 describes our feature extraction methods and different types of analys that we use. This section also shows the block diagram of the training and testing phases. Section~3 explains the use of transfer learning by importing a part of VGG16 network and Section~4 reports the experimental results and comparisons with state-of-the-art methods. Finally, Section~5 concludes the paper.

\section{Appearance-Motion and Motion-Only Analyses}
To detect abnormalities in videos, researchers often analyze either appearance features or motion features. However, processing both appearance and motion features is usually necessary because this combination may include additional important information.

\begin{figure*}
\centering
\includegraphics[width=17cm]{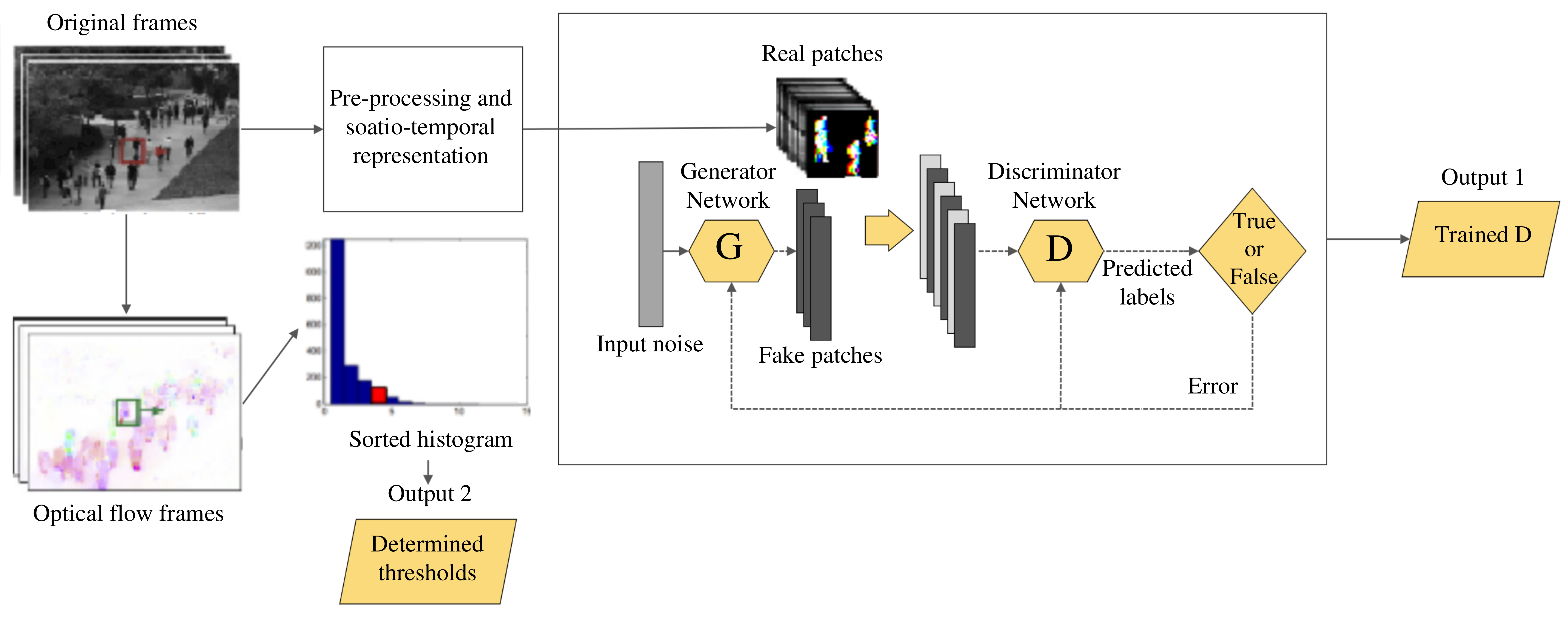}
\caption{Training phase of the proposed method.}
\label {fig1}
\end{figure*}

In this work, we take advantages of both appearance and motion features. In the proposed model, the input which is a sequence of original frames moves in two different paths. One path (straight path shown in Fig.~\ref{fig1}) is appearance-motion analysis which extracts 3D spatio-temporal features. Another path (downward path shown in Fig.~\ref{fig1}) is motion-only analysis which extracts abnormal directions or speeds using optical flow images.

\subsection{Pre-processing of Input Images}
\begin{figure}
\centering
\includegraphics[width=8cm]{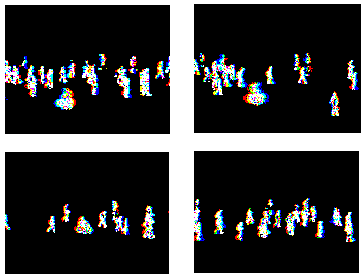}
\caption{Spatio-temporal representation of three frames.}
\label {fig2}
\end{figure}

Pre-processing of the data is typically the first step to be taken prior to using any learning model. In this step, certain image processing tasks are applied to the frames to resolve their appearance challenges and prepare them to enter the learning model. Examples of the tasks that can be done in the pre-processing step are histogram equalization, foreground extraction, edge detection of foreground objects, obtaining spatio-temporal representations, and patch extraction. A 3D structure is formed to jointly consider the motion features and the appearance features. Fig.~\ref{fig2} shows the spatio-temporal representation of three frames chosen by picking every other frame in a set of consecutive frames.\footnote{Using every other frame instead of using three consecutive frames better shows the movement of the objects.} The spatio-temporal representation of frame number $t$ is
\begin{equation}
D_t=<I_t,I_\text{t-2},I_\text{t-4}>,
%\label {star}
\end{equation}
where $I\textsubscript{t}$ is the edge image of frame number $t$.

\subsection{Patch Extraction From Appearance-Motion Representation}
\begin{figure}
\centering
\includegraphics[width=8cm]{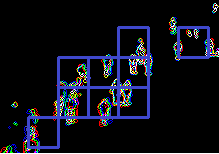}
\caption{Patch selection from spatio-temporal representation.}
\label {fig3}
\end{figure}

Frame patches are determined by adding a grid overlay to the appearance-motion representation of the frames. Among all patches, those that have at lease a minimum amount of foreground pixels are chosen to avoid noises and useless data. These patches are gathered from video frames and inputted to the network for learning and testing processes. Fig.~\ref{fig3} shows the output of the patch selection unit.

\subsection{Objects Movement Analysis}
In this work we focus on the hue and intensity of the output of Gunner-Farneback optical flow algorithm~\cite{farneback2003two}. The hue shows the path direction and the intensity shows the speed of moving objects in a video sequence. 

\begin{figure*}
\centering
\includegraphics[width=17cm]{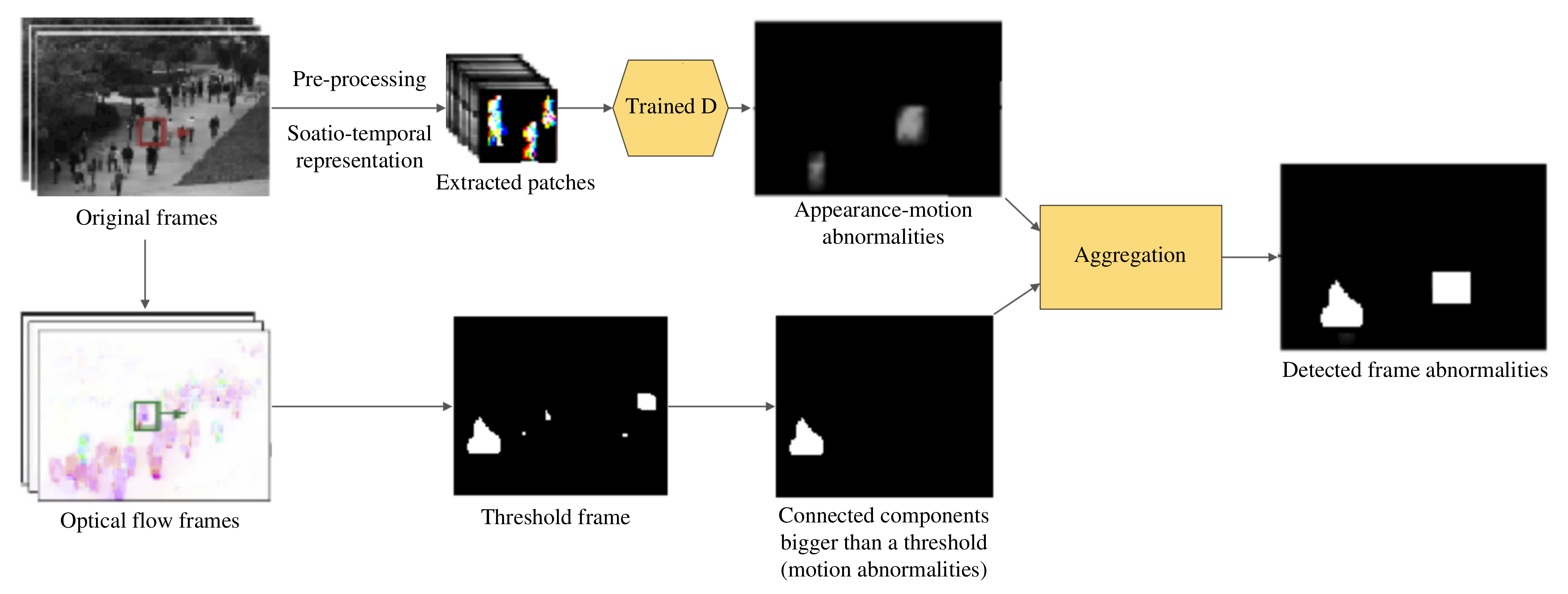}
\caption{Testing  phase of the proposed method.}
\label {fig4}
\end{figure*}

Fig~\ref{fig4} shows the block diagram of the testing phase of the proposed method. Sorted histograms of direction and speed intensities are calculated in all video frames. Motion abnormalities are then detected based on the sorted histograms. The first $5\%$ of hue/intensity values with the lowest frequencies in the sorted histogram are considered as abnormal.
As a result, motion abnormalities are obtained from this step. The result of this analysis, together with that of the appearance-motion analysis can effectively detect and locate abnormal patches in the video. 

\section{Transfer Learning to Benefit From VGG16}
Data acquisition is one of the main challenges of every machine learning problem. Also, most of the machine learning methods- especially deep learning--based methods- are very time consuming. One of the potential solutions is to transfer a pre-trained model with a specific data domain to a related but different domain without the need for re-learning or providing new datasets. For instance, a model that learned to detect cars in a video sequence can detect unseen trucks without re-learning procedure. This concept is referred to “transfer learning” that introduced in~\cite{pratt1993discriminability}.% This concept can be considered as proverbs in a language that can help clarify a concept by a statement in a different domain of speech~\cite{pan2009survey}.%

In this work we use GAN that introduced in~\cite{goodfellow2014generative} to learn the normal data distribution using the normal appearance-motion patches. In each step, the generator outputs patches and gives them to a discriminator that uses a pre-trained VGG16 network. The discriminator then guesses the originality of its input patches. The final error of the discriminator is calculated based on the accuracy of its detections and returned to both the generator and the discriminator. Thus, the generator improves its fake generation accuracy and the discriminator improves its ability to distinguish between real and fake images.

\begin{figure}
\centering
\includegraphics[width=8cm]{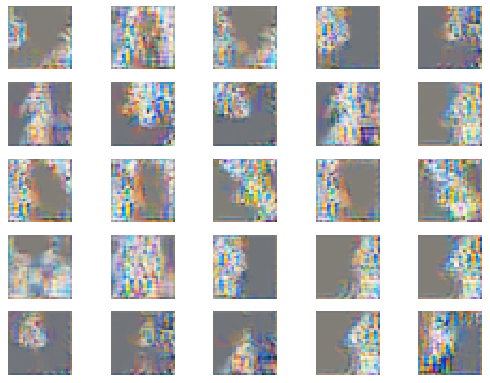}
\caption{The generator outputs after $16000$ iterations.}
\label {fig5}
\end{figure}

In this work, the first six layers of a pre-trained VGG16 network are used to obviate the need for training more than $10$~million learnable parameters and gathering a large amount of various data. Fig.~\ref{fig5} shows the results of this network after $16000$ iterations. At first, the generator generates random noises. However, after $16000$ iterations the generated patches become very similar to pedestrians in the real patches. This means that the generator outputs are improving. 

\section{Experimental results}
The efficiency of the proposed method is evaluated with AUC and EER criteria. The comparison between the proposed method and the state of the art shows that the proposed method is very effective in detecting and locating abnormalities in the videos.

Python programming language and Keras module are used to implement the GAN subnetworks. The computer used in this experiment has NVIDIA GeForce~$1050$~GPU, $32$~GB of RAM, and an Intel Core~i$7$ CPU running at $3.1$~GHz. To evaluate the model, $\text{roc\_curve}$ function of Python sklearn module is used to calculate the area under the ROC curve. The generator has four convolutional layers and the discriminator has two parts: The first part is a portion of VGG16 from the input to the pooling layer of the fourth block. This part is constant and unlearnable. The second part consists of two fully-connected learnable layers which are concatenated to the first part.

\subsection{The utilized datasets: $\text{UCSD Peds1}$ and $\text{UCSD Peds2}$}
$\text{UCSD~Peds}$ dataset includes two subsets: $\text{UCSD Peds1}$ and $\text{UCSD Peds2}$. Both show a crowded pedestrian zone where bikers, skaters, and carts are considered abnormal entities. This dataset has 50 training and 48 testing video samples. The difference between the two subsets is their frame size and the angle of the camera. In $\text{UCSD Peds1}$, the camera is placed on a high altitude. $\text{UCSD Peds1}$ has perspective distortion and a resolution of 238*158 pixels. In $\text{UCSD Peds2}$ the camera is placed on a lower altitude. It has a resolution of 360*240 pixels and NO perspective distortion.

\subsection{Model Evaluation Criteria}
There are two types of evaluation for AED models and approaches: Frame-level evaluation and pixel-level evaluation. In frame-level evaluation, the model accuracy is calculated based on the abnormal frame detection without localizing the abnormality. On the contrary, pixel-level evaluation localizes abnormalities and checks the model output with the pixels of ground-truth images. If a model can correctly detect at least 40\% of abnormal pixels, its detection is considered true.

The area under the ROC curve which is drawn based on true positive rate (TPR) and false positive rate (FPR) is a common criterion to evaluate and compare the performance and accuracy of different models. TPR is calculated as
\begin{equation}
\text{TPR}=\frac{TP}{P}=\frac{TP}{TP+FN},
\label{eqTPR}
\end{equation}
where $TP$ is the number of frames that are correctly detected as abnormal frames and $FN$ is the number of frames that are incorrectly considered as normal frames. Equation~(\ref{eqTPR}) can also be used in the pixel-level mode but instead of the number of frames, the number of pixels should be counted. FPR is calculated as

\begin{equation}
\text{FPR}=\frac{FP}{N}=\frac{FP}{FP+TN},
\end{equation}
where $FP$ is the number of frames that are incorrectly detected as abnormal frames and $TN$ is the number of frames that are correctly considered as normal frames.

\begin{figure}
\centering
\includegraphics[width=7cm]{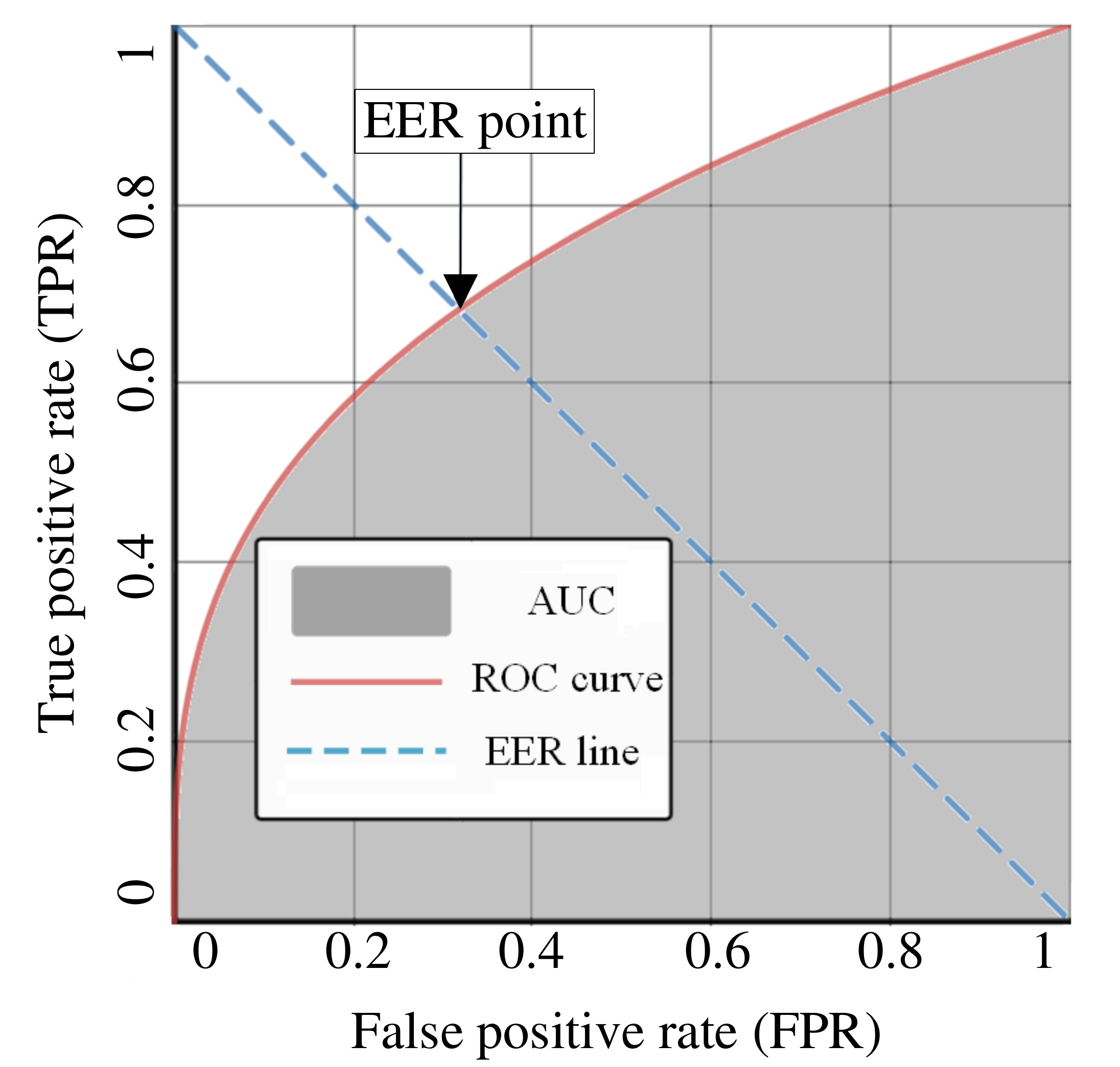}
\caption{EER and ROC.}
\label {fig6}
\end{figure}

Fig.~\ref{fig6} shows the Equal error rate (EER) which is a point in the ROC at the intersection of the curve and a line that traverses from (0,1) and (1,0). False negative rate (FNR) is the proportion of positives which yield negative test outcomes that calculated as:
\begin{equation}
\text{FNR}=\frac{FN}{P}=\frac{FN}{TP+FN},
\end{equation}
At EER, FPR and FNR are equal. A lower EER shows that the algorithm is more accurate with less error. 
Another important criterion is time complexity. An algorithm is more attractive to use in different applications if  its overall execution time is sufficiently short.

\subsection{Results with $\text{UCSD Peds1}$ and $\text{UCSD Peds2}$}

\begin{figure}
\centering
\includegraphics[width=8cm]{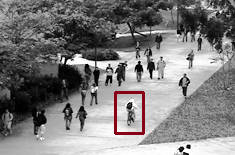}
\caption{Detected abnormality with $\text{UCSD Peds1}$.}
\label {fig7}
\end{figure}

After training the model with the training data, the model is tested with testing data and the system is evaluated with the above-mentioned criteria by using the ground-truth of the dataset. Fig.~\ref{fig7} shows an instant of testing the system when it is detecting a biker as an abnormality. 

\begin{figure}
\centering
\includegraphics[width=8cm]{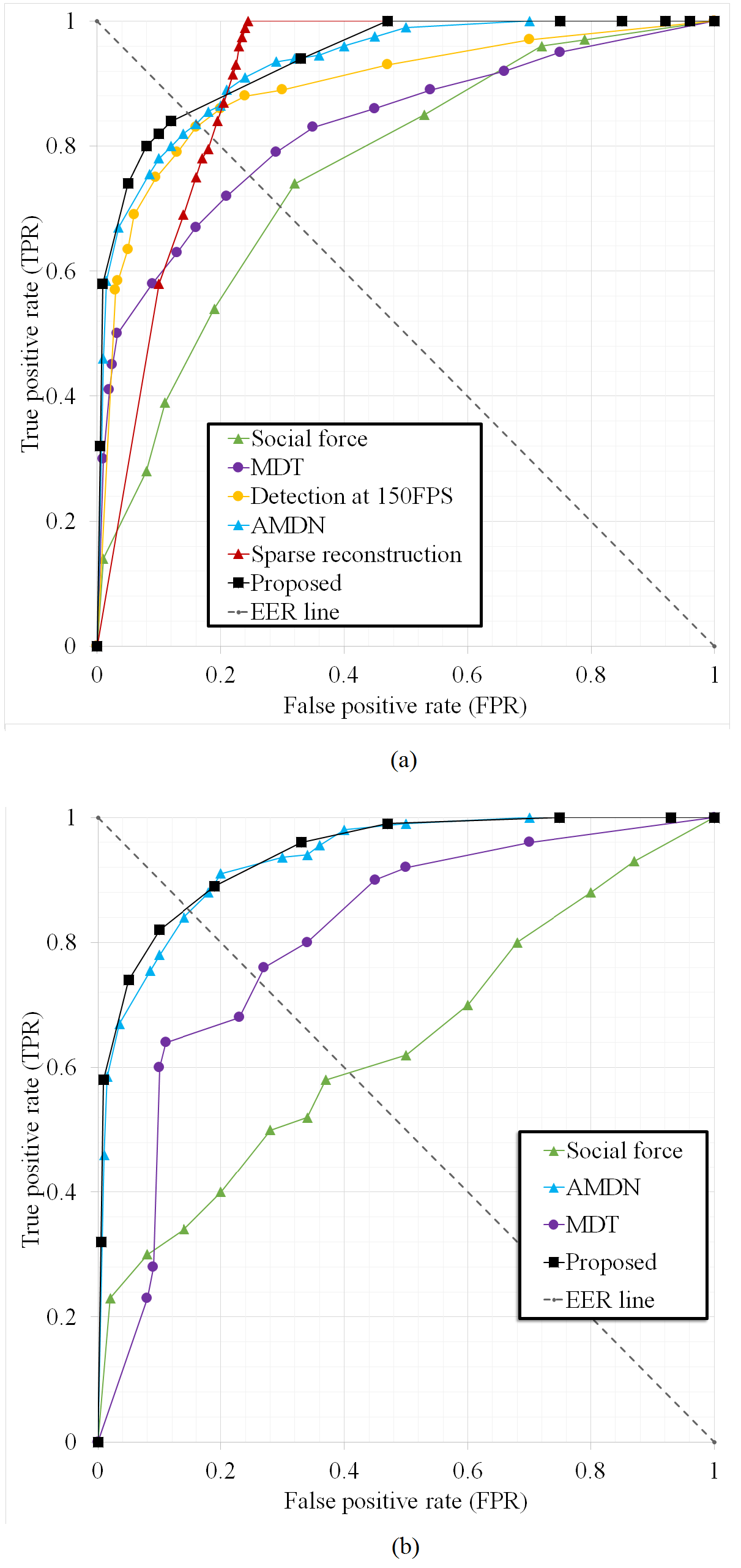}
\caption{Frame-level ROC comparison of the proposed method with state of the art. (a) when $\text{UCSD Peds1}$ is used and (b) when $\text{UCSD Peds2}$ is used.}
\label {fig8}
\end{figure}

\begin{figure}
\centering
\includegraphics[width=8cm]{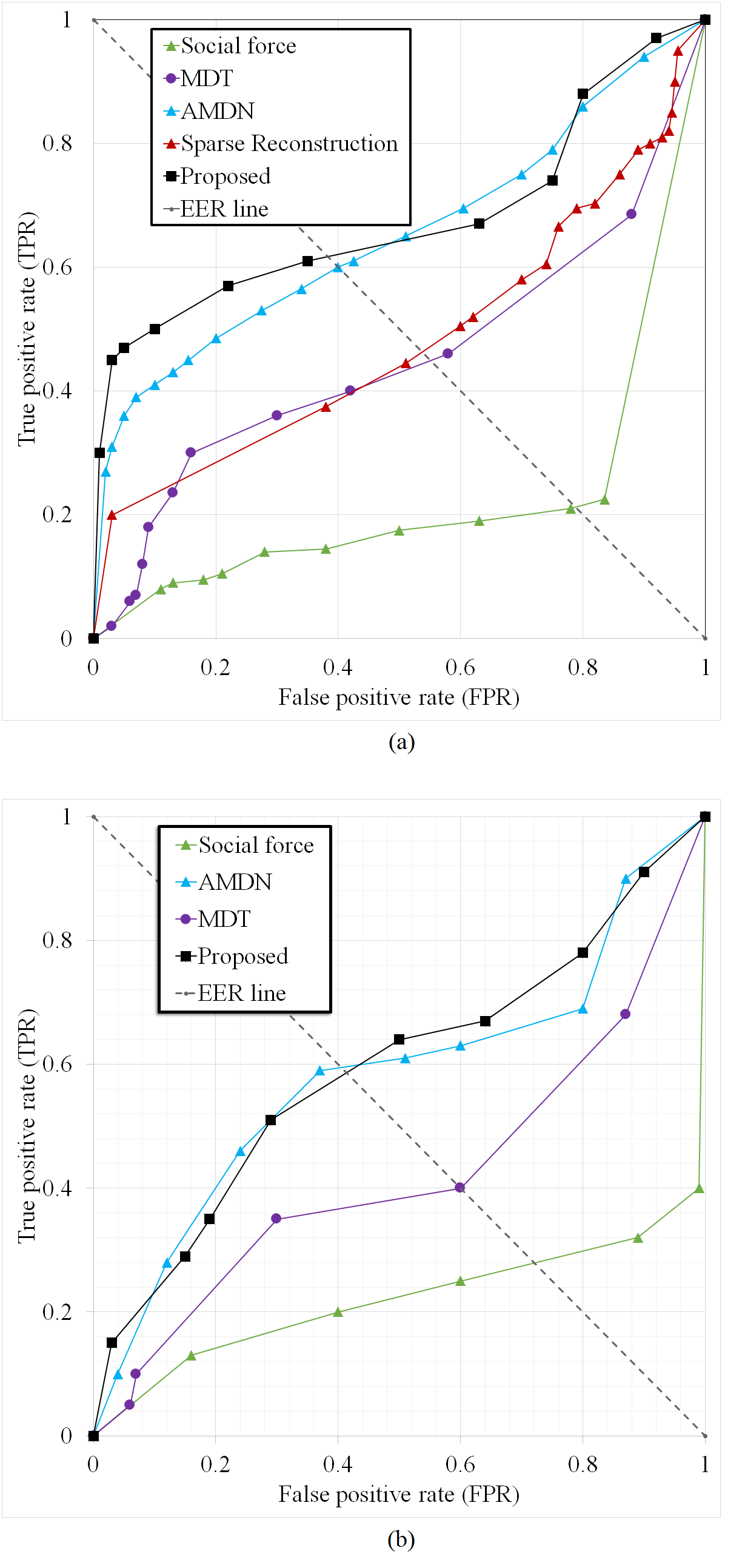}
\caption{Frame-level ROC comparison of the proposed method with state of the art. (a) when $\text{UCSD Peds1}$ is used and (b) when $\text{UCSD Peds2}$ is used.}
\label {fig9}
\end{figure}

Figs.~\ref{fig8}(a) and \ref{fig9}(a) show that compared to the state-of-the-art AED methods, the proposed method is more effective in both frame-level and pixel-level evaluations as it has a smaller EER and a larger AUC.\footnote{ROC curves of the previous methods are derived from~\cite{ravanbakhsh2017abnormal}.}

Fig.~\ref{fig9}(a) shows that the overall accuracy and AUC of all methods in the pixel-level evaluation are less than those in the frame-level evaluation. This is because the pixel-level evaluation is a  more stringent evaluation: In addition to evaluating the ability of an AED method in detecting the abnormalities of the frames, it also evaluates the ability of the AED method in locating the abnormalities. This results in larger EER and smaller AUC values as compared to frame-level evaluation. 
%The location of an abnormality is important in this mode. {\color{red}This is because the pixels are individually considered in determining the correctness of the prediction. However, frame-level evaluation considers a whole frames prediction. In frame-level mode, the prediction of a model is the abnormality of the whole frame. Thus, the location of abnormality is not considered.} 
Table~\ref{table1} lists the values of AUC and EER of each method and further validates the effectiveness of the proposed method.

The main reason for the superiority of the proposed method is that is uses both appearance and motion analyses. With $\text{UCSD Peds1}$, detection of motion abnormalities is important: Skaters, carts, and bikers usually move faster than pedestrians. Also with more challenging abnormalities such as low-speed bikers, appearance analysis of the proposed method is useful in achieving effective results. The proposed method also has remarkable results with $\text{UCSD Peds2}$ dataset. Table~\ref{table2} shows the superiority of the proposed method over the state-of-the-art methods using EER comparison.

\begin{table}[ht]
  \begin{center}
    \caption{EER and AUC comparisons with UCSD Peds1.}
    \label{table1}
    \begin{tabular}{lcccc} % <-- Changed to S here.
    \hline
      \multirow{2}{*}{\textbf{Method}} &\hspace{-0.3cm}
      \textbf{Frame-level} &\hspace{-0.3cm}
      \textbf{Frame-level} & \hspace{-0.3cm}
      \textbf{Pixel-level} & \hspace{-0.3cm}
      \textbf{Pixel-level} \\
      \textbf{} & \textbf{EER} &\hspace{-0.3cm} \textbf{AUC} &\hspace{-0.3cm} \textbf{EER} &\hspace{-0.3cm} \textbf{AUC} \\
      \hline
      MDT [7] & 25\% & 81\% & 58\% & 44\% \\
      Social force [8] & 31\% & 68\% & 79\% & 20\% \\
      AMDN [3] & 16\% & 92\% & 40\% & 67\% \\
      Proposed & \textbf{14\%} & \textbf{93\%} & \textbf{36\%} & \textbf{73\%\vspace{0.1 cm}} \\
      \hline
    \end{tabular}
  \end{center}
\end{table}

\iffalse
\begin{table}[ht]
  \begin{center}
    \caption{Frame-level and pixel-level EER comparisons with $\text{UCSD Peds2}$.}
    \label{table2}
    \begin{tabular}{ccc} % <-- Changed to S here.
    \hline
      \multirow{2}{*}{\textbf{Method}} &
      \textbf{Frame-level} &
      %\textbf{Frame-level} & 
      \textbf{Pixel-level} %& 
      %\textbf{Pixel-level}
      \\
      \textbf{} & 
      \textbf{EER} & 
      %\textbf{AUC} & 
      \textbf{EER} %& 
      %\textbf{AUC}
      \\
      \hline
      MDT [7] & 24\% & 54\% \\ % & 58.0\% & 44.1\% \\
      Social force [8] & 42\% & 80\% \\ %& 79.0\% & 19.7\% \\
      AMDN [3] & 16\% & 42\% \\ %& 40.1\% & 67.2\% \\
      Proposed & \textbf{15\%} & \textbf{17\%} \\ %& \textbf{36.2\%} & \textbf{73.1\%} \\
      \hline
    \end{tabular}
  \end{center}
\end{table}
\fi

\begin{table}[ht]
  \begin{center}
    \caption{Frame-level and pixel-level EER comparisons with $\text{UCSD Peds2}$.}
    \label{table2}
    \begin{tabular}{lcc} % <-- Changed to S here.
    \hline
    \textbf{Method} & \hspace{0.3cm} \textbf{Frame-level EER} & \hspace{0.3cm} \textbf{Pixel-level EER} \\ % & 58.0\% & 44.1\% \\
      
      \hline
      MDT [7] & 24\% & 54\% \\ % & 58.0\% & 44.1\% \\
      Social force [8] & 42\% & 80\% \\ %& 79.0\% & 19.7\% \\
      AMDN [3] & 16\% & 42\% \\ %& 40.1\% & 67.2\% \\
      Proposed & \textbf{15\%} & \textbf{17\%} \\ %& \textbf{36.2\%} & \textbf{73.1\%} \\
      \hline
    \end{tabular}
  \end{center}
\end{table}

\subsection{Time Analysis}

For GAN to be sufficiently accurate, its learning process requires a large number of repetitions. The time complexity for both training and testing phases is one of the major problems in most deep neural networks including GAN. However, we decrease the time complexity of the GAN by using transfer learning. In this work, VGG16 network is used to extract the appearance features.
VGG16 is a well-known pre-trained network which is trained with ImageNet dataset. ImageNet dataset contains more than 14 million images from more than 21800 categories. VGG16 is trained by a large number of various data. Thus, its feature extractor is general enough and we can use its first layers as the feature extractor of our method. To show the effectiveness of the proposed method, we noted the execution times of different parts of our method. Major execution times that should be analyzed in the learning process of the proposed method are pre-processing time, optical flow extraction time, 3D representation time, and classification time.  Table~\ref{table3} shows these execution times for the proposed method. 

\begin{table}[ht]
  \begin{center}
    \caption{
    Execution time of each task for one frame.}
    \label{table3}
    \begin{tabular}{ccccc} % <-- Changed to S here.
    \hline
    Pre- & Optical & 3D & Classification & Total
%      \multirow{2}{*}{\textbf{Method}} &
      \\
      processing & flow & representation & &
      \\
      & extraction & & &
      \\
      \hline
        0.001~s & 0.020~s & 0.001~s & 0.290~s & 0.312~s
        \\
      \hline
    \end{tabular}
  \end{center}
\end{table}

\section{Conclusion}
In this paper, a new GAN-based method is proposed to address AED problem. This method uses both appearance-motion and motion-only representations of the input data. Therefore, it effectively detects various abnormalities in shape, skeleton, speed, and direction. The proposed method transfers the knowledge of a pre-trained CNN (VGG16) to its discriminator CNN to solve this unsupervised problem. This knowledge transfer makes the training phase of the proposed method highly efficient. Experimental case studies are carried out to compare the proposed method with the state of the art. The experiments are based on $\text{UCSD Peds1}$ and $\text{UCSD Peds2}$ datasets.

The results show the effectiveness of the proposed method as it has a lower EER and higher AUC as compared to other methods. In addition to the performance studies, time complexity study is also carried out which shows that the proposed method is sufficiently time effective. With the utilized personal computer, the proposed method can detect and locate abnormalities of a frame in a short time. By using GPU oriented codes and with the advancements of the computers, the proposed method will also be applicable for real-time AED with higher video frame rates (e.g., 30+ fps).

\bibliography{paper}
\bibliographystyle{IEEEtran}

\pagestyle{plain}
\end{document}